\documentclass[10pt,twocolumn,letterpaper]{article}

\usepackage[pagenumbers]{wacv} 

\usepackage{booktabs}
\usepackage{multirow}
\usepackage{colortbl}
\usepackage{graphicx}
\usepackage{caption}
\usepackage{subcaption}
\usepackage{placeins}

\definecolor{wacvblue}{rgb}{0.21,0.49,0.74}
\usepackage[pagebackref,breaklinks,colorlinks,allcolors=wacvblue]{hyperref}

\def\wacvPaperID{27} 
\def\confName{WACV}
\def\confYear{2027}

\title{
Gradient-Guided Orientation Learning for\\
Weakly Supervised Oriented Object Detection
}

\author{
Junyong Jang$^{*}$ \quad
Juheon Park$^{*}$ \quad
Minyoung Jeon \quad
Jaehyup Lee$^{\dagger}$\\
Kyungpook National University, South Korea\\
{\tt\small bogwang6@knu.ac.kr, juheona@knu.ac.kr, mym0314@knu.ac.kr, jaehyuplee@knu.ac.kr}
}

\begin{document}

\twocolumn[{
\maketitle

\begin{center}
    \centering
    \includegraphics[width=0.97\textwidth]{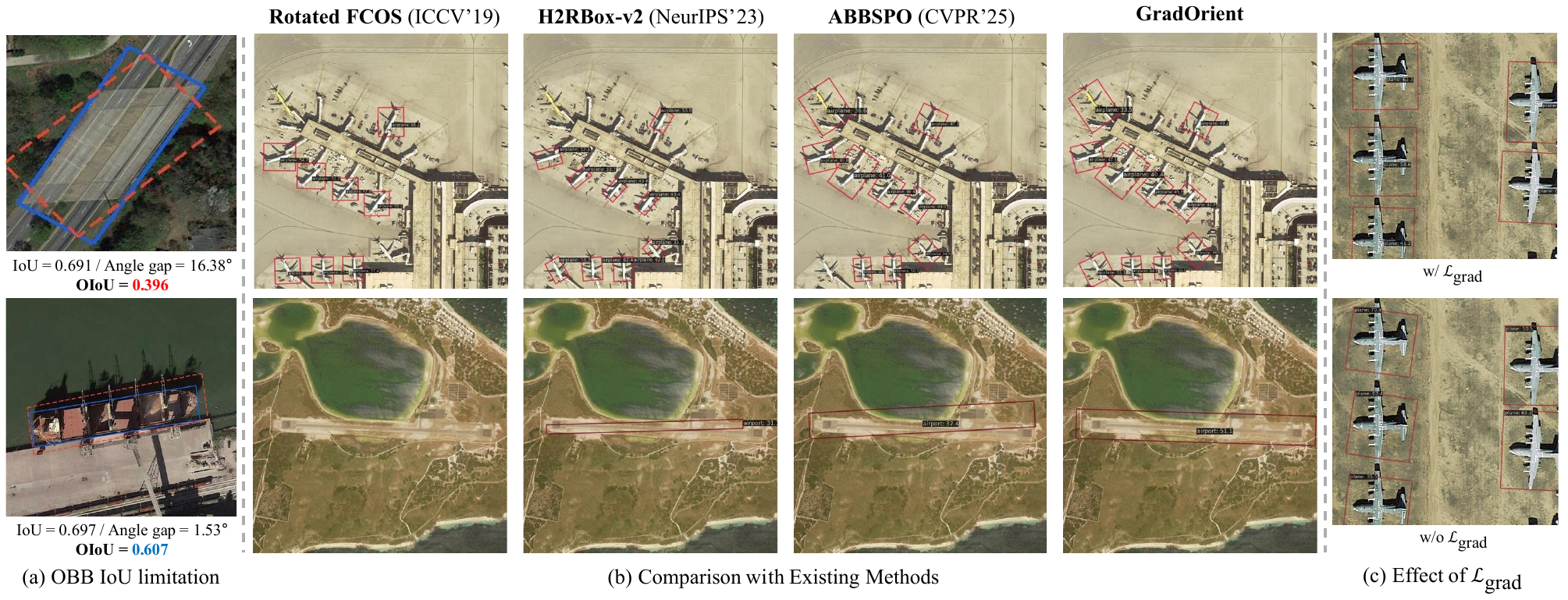}
    \vspace{-0.2cm}
    \captionof{figure}{Qualitative comparison of detection results.
(a) Similar OBB IoU values can correspond to substantially different angle errors.
(b) GradOrient provides more accurate detections than existing methods in challenging scenes containing densely distributed or large elongated objects.
(c) $\mathcal{L}_{\text{grad}}$ improves fine-grained angle estimation and oriented bounding-box alignment.}
    \label{fig:teaser}
\end{center}
}]

\begingroup
\renewcommand{\thefootnote}{}
\footnotetext{
    \textsuperscript{*}Equal contribution.
    \qquad
    \textsuperscript{$\dagger$}Corresponding author.
}
\endgroup

\begin{abstract}
Weakly supervised oriented object detection (WS-OOD) aims to reduce the annotation cost of rotated bounding boxes by leveraging cheaper annotations such as horizontal boxes or points. However, these weak annotations provide no explicit orientation information, forcing existing methods to infer object angles through indirect geometric consistency or equivariant constraints. Such indirect supervision often becomes ambiguous and unstable under real-world appearance variations, including illumination changes, color shifts, shadows, and background clutter. In this paper, we present a simple yet effective gradient-guided orientation learning strategy for WS-OOD. Motivated by the observation that the input image itself contains useful orientation priors, our method extracts dominant gradient-based orientation cues from object regions and converts them into pseudo-angle labels for auxiliary supervision, enabling the detector to learn orientation-aware representations without requiring rotated bounding box annotations or extra learnable parameters. Furthermore, since AP can mask substantial angular errors when spatial IoU remains high, we introduce an orientation-aware evaluation metric that incorporates angular discrepancies into the standard AP protocol. Extensive experiments on weakly supervised oriented detection benchmarks demonstrate that the proposed pseudo-angle supervision improves both detection performance and orientation estimation quality, while the proposed metric provides a more faithful assessment of angle prediction than conventional AP alone.
\end{abstract}

\section{Introduction}
\label{sec:intro}

Oriented object detection (OOD) aims to localize objects with rotated bounding boxes (RBoxes), providing both object location and orientation information. This task is particularly important in remote sensing, where objects such as ships, aircraft, vehicles, bridges, and harbors frequently appear with arbitrary orientations. Compared with annotating horizontal bounding boxes (HBoxes), however, annotating RBoxes requires additional angle information and is considerably more labor-intensive \cite{yu2024point2rbox}. This annotation burden has motivated weakly supervised oriented object detection (WS-OOD), which learns oriented detectors from cheaper annotations such as points or HBoxes. Despite their cost-effectiveness, these weak annotations provide localization cues but do not explicitly specify object orientation, making accurate angle recovery inherently ambiguous.

To address the absence of direct angle labels, existing WS-OOD methods \cite{lee2025abbspo, yang2022h2rbox, yu2023h2rbox, luo2024pointobb, yu2024point2rbox, yu2025point2rbox, eie} mainly infer object orientations through transformation consistency, equivariant constraints, or geometric priors. These methods demonstrate that useful orientation cues can be obtained by enforcing prediction consistency across original, rotated, and flipped views. However, such cues remain indirect, since multiple orientations may satisfy the same weak supervision constraint. This ambiguity makes precise angle estimation difficult, especially under real-world appearance variations such as illumination changes, color shifts, shadows, occlusions, and background clutter. Therefore, a practical WS-OOD framework requires an additional orientation-aware cue that complements geometric consistency without increasing annotation cost.

In weakly supervised learning, where the provided labels are intentionally incomplete, the design of supervision plays an essential role. Nevertheless, one important source of orientation information has been relatively underexplored: the input image itself. Remote sensing objects often exhibit elongated structures, boundary edges, and local gradient patterns that are strongly correlated with their physical orientations. Unlike external labels or manually defined class priors, these visual cues are already available in every training image and can serve as object-specific orientation priors at no additional annotation cost. This observation raises a practical question: can we convert image-intrinsic orientation cues into explicit angle supervision for WS-OOD?

Motivated by this question, we propose \textbf{GradOrient}, a simple yet effective gradient-guided orientation learning strategy for weakly supervised oriented object detection. Inspired by histograms of oriented gradients (HOG) \cite{dalal2005histograms}, GradOrient extracts dominant orientation cues from weakly annotated object regions and converts them into pseudo-angle labels for auxiliary supervision. Specifically, we estimate object-specific dominant directions using magnitude-weighted local gradient orientation histograms and assign confidence scores to suppress unreliable pseudo-angle signals. Unlike purely geometric formulations, the proposed supervision directly encourages the detector to learn orientation-aware representations from image evidence. Since GradOrient uses only the input image and weak annotations, it does not require RBox labels, additional manual annotation, or extra learnable parameters. Although we validate our method in the HBox-to-RBox setting, the proposed strategy can be incorporated into various WS-OOD frameworks as a plug-in supervision signal.

In addition to training supervision, we identify an important limitation in the evaluation protocol of existing OOD and WS-OOD methods. Current studies predominantly rely on average precision (AP), which evaluates detection quality based on classification confidence and spatial overlap. While AP is widely used for object detection, it does not explicitly measure orientation accuracy. In many cases, predictions with substantially different angles can still achieve sufficiently high IoU values and be counted as true positives \cite{eie, fu2025bghr, yang2022detecting, yang2022kfiou}. As shown in Fig. \ref{fig:teaser}(a), this issue becomes more severe for objects whose width and height are similar, where large angular errors may have only a limited effect on IoU. Since object orientation is critical for downstream remote sensing tasks such as maritime monitoring, aircraft pose interpretation, and infrastructure inspection, AP alone cannot fully reflect the quality of angle estimation.

To provide a more faithful assessment of orientation prediction quality, we introduce an orientation-aware evaluation metric that explicitly penalizes angular discrepancies between predicted and target boxes. The proposed metric preserves the conventional AP evaluation pipeline while incorporating an angular penalty into the matching criterion, enabling complementary analysis of both localization and orientation quality. We further evaluate existing OOD methods with the proposed metric, showing that conventional AP can mask substantial angle errors and that our gradient-guided supervision improves orientation estimation beyond standard IoU-based evaluation. Our main contributions are summarized as follows:

\begin{itemize}
\item We propose \textbf{GradOrient}, a gradient-guided pseudo-angle supervision strategy that improves WS-OOD without requiring RBox annotations.

\item We introduce an image-intrinsic orientation prior that extracts dominant gradient directions from input data and can be plugged into existing WS-OOD frameworks.

\item We present an orientation-aware evaluation metric that complements conventional AP by explicitly penalizing angular discrepancies in oriented detection.
\end{itemize}

\section{Related Work}
\label{sec:related_work}

\subsection{RBox-supervised Oriented Object Detection}

Early oriented object detection (OOD) methods extended conventional object detectors to predict rotated bounding boxes (RBoxes), including Rotated RetinaNet~\cite{lin2017focal}, Rotated FCOS~\cite{tian2019fcos}, RoI Transformer~\cite{ding2019learning}, Gliding Vertex~\cite{xu2020gliding}, Oriented R-CNN~\cite{xie2021oriented}, and Oriented RepPoints~\cite{li2022oriented}. Subsequent studies have further improved orientation modeling and localization quality. R$^3$Det~\cite{yang2021r3det} employs progressive regression with feature reconstruction, S$^2$A-Net~\cite{han2021align} aligns features according to object orientations, and ReDet~\cite{han2021redet} introduces rotation-equivariant representations. In addition, Gaussian-based formulations such as GWD~\cite{yang2021rethinking}, KLD~\cite{yang2021learning}, and KFIoU~\cite{yang2022kfiou} alleviate the angle boundary discontinuity problem through alternative box representations and loss functions. Despite these advances, most OOD methods rely on fully annotated RBoxes, which are considerably more expensive to obtain than horizontal annotations. This annotation burden has motivated weakly supervised OOD.

\subsection{Weakly Supervised Oriented Object Detection}

Weakly supervised oriented object detection aims to reduce the annotation cost of RBoxes by learning from weaker supervision, such as point annotations or horizontal bounding boxes (HBoxes). Depending on the supervision type, existing methods can be broadly categorized into point-supervised and HBox-supervised approaches. More recently, Wholly-WOOD~\cite{yu2025wholly} provides a unified framework for both weakly and fully supervised OOD, while PWOOD~\cite{liu2026partial} further explores a partially weakly supervised setting with mixed supervision.

\noindent\textbf{Point-supervised Oriented Object Detection.}
Point-supervised OOD~\cite{cao2023p2rbox, he2023learning, yu2025point2rbox, zhang2025pointobb, zhang2024pmho} reduces annotation cost by replacing bounding boxes with single-point labels. Point2RBox~\cite{yu2024point2rbox} introduces an end-to-end framework that combines synthetic visual pattern knowledge with transformation self-supervision. PointOBB~\cite{luo2024pointobb} generates pseudo RBoxes through scale-sensitive consistency learning and multiple instance learning, while PointOBB-v2~\cite{ren2025pointobb} further improves pseudo-label quality using class probability maps and geometric priors. These methods show that oriented detection can be learned from sparse annotations, but accurate angle estimation still depends heavily on indirectly generated pseudo labels and geometric assumptions.

\noindent\textbf{HBox-supervised Oriented Object Detection.}
HBox-supervised OOD~\cite{lu2024afws, sun2021oriented, zhu2023knowledge, fu2025bghr} has attracted increasing attention because HBoxes provide stronger localization cues than points while remaining much cheaper to annotate than RBoxes. H2RBox~\cite{yang2022h2rbox} learns object orientations through rotation consistency, and H2RBox-v2~\cite{yu2023h2rbox} further exploits symmetry and flip consistency. EIE-Det~\cite{eie} enhances equivariance-based learning through instance-level orientation constraints, while ABBSPO~\cite{lee2025abbspo} leverages object symmetry as an additional cue for improving orientation estimation. Although these methods are effective, they mainly infer orientations through transformed-view consistency or predefined geometric priors. As a result, the supervision for absolute angle estimation remains indirect.

\noindent\textbf{Position of Our Work.}
Unlike prior HBox-supervised methods~\cite{yang2022h2rbox,yu2023h2rbox,lee2025abbspo,eie}, our work exploits image-intrinsic gradient evidence as a practical orientation prior. We convert gradient orientation statistics from weakly annotated object regions into explicit pseudo-angle supervision, complementing existing geometric consistency constraints without requiring RBox annotations. Beyond training, we also introduce an orientation-aware evaluation metric that reveals angle errors overlooked by conventional AP. Together, our method provides both a simple supervision signal and a more faithful evaluation protocol for practical WS-OOD.
\section{Method}
\label{sec:method}
\subsection{Overall Pipeline}
\label{subsec:overall_pipeline}
\begin{figure*}[t]
    \centering
    \includegraphics[width=1.0\textwidth]{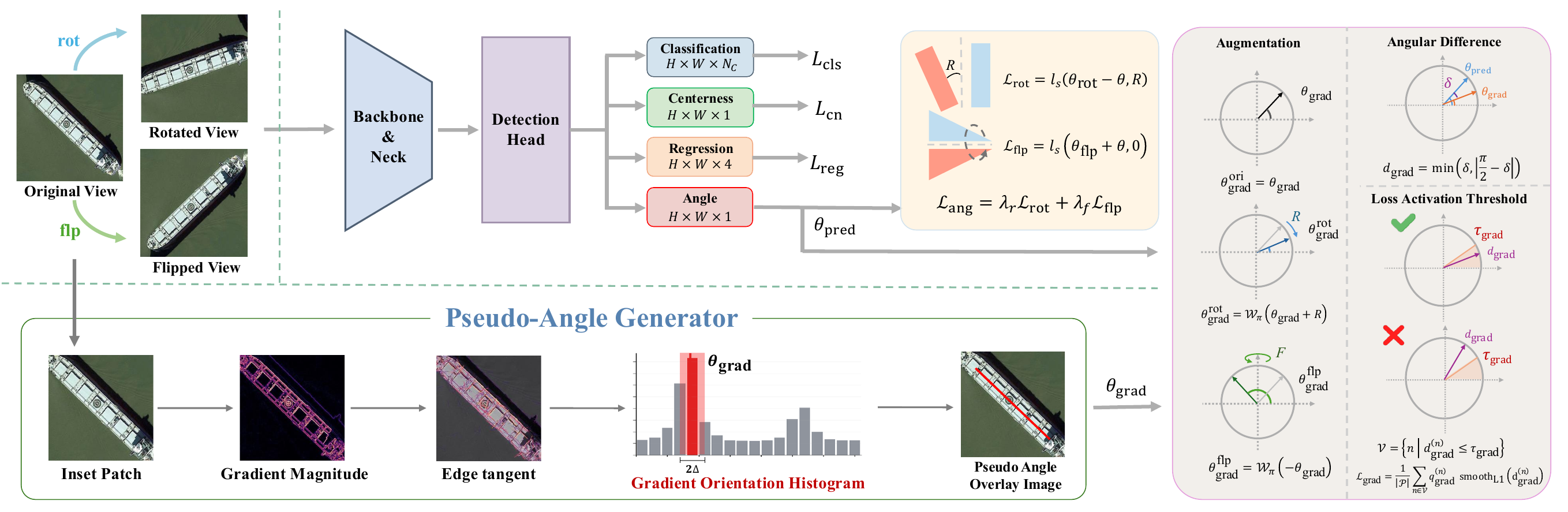}
    \vspace{-0.5cm}
    \caption{Overall pipeline of GradOrient. A pseudo-angle generator estimates object-specific orientation cues from magnitude-weighted gradient orientation histograms within weakly annotated regions. The resulting pseudo angles are transformed according to rotated and flipped views and used for gradient-guided orientation supervision through threshold-based loss activation.}
    \label{fig:my_image_label}
    \vspace{-0.3cm}
\end{figure*}

The overall pipeline of the proposed framework is illustrated in \cref{fig:my_image_label}. We adopt H2RBox~\cite{yang2022h2rbox}, H2RBox-v2~\cite{yu2023h2rbox}, and ABBSPO~\cite{lee2025abbspo} as baseline architectures. The input image and its rotated and flipped views are processed through a backbone and a Feature Pyramid Network (FPN). A shared detection head then predicts class probabilities, centerness, and spatial parameters of the RBox.

On top of this WS-OOD framework, we introduce a gradient-guided orientation learning module that provides pseudo-angle supervision. The module estimates dominant orientation cues from image gradients within each annotated HBox and uses them as auxiliary targets for angle learning. This allows the detector to exploit image-intrinsic orientation priors without requiring RBox annotations.

\subsection{Pseudo-Angle Generator}
To obtain a gradient-guided pseudo-angle $\theta_{\text{grad}}$ for orientation learning, we use local image gradients within an inset region of each annotated HBox in the augmented training input. We first apply Gaussian smoothing to each inset HBox patch and compute pixel-level gradient responses. Then, we retain only the top-$k\%$ pixel-level gradient responses ranked by gradient magnitude within each inset HBox patch. The retained gradient orientations are converted into edge-tangent orientations and accumulated into a magnitude-weighted orientation histogram~\cite{lowe2004distinctive, dalal2005histograms}.

For the $i$-th retained pixel, let $G_x(i)$ and $G_y(i)$ denote the horizontal and vertical image-gradient components, respectively. We define its gradient orientation $\alpha_i$ and magnitude $m_i$ as $\alpha_i=\operatorname{atan2}(G_y(i),G_x(i))$ and $m_i=\sqrt{G_x(i)^2+G_y(i)^2}$.

Since \(\alpha_i\) corresponds to the edge-normal direction, we convert it into the edge-tangent orientation as:
\begin{equation}
\phi_i = \left( \alpha_i+\frac{\pi}{2} \right) \bmod\pi .
\end{equation}

We then construct a magnitude-weighted orientation histogram using the retained pixels. Let $\mathcal{R}$ denote the set of retained pixels, $B$ be the number of orientation bins, and $\Delta=\pi/B$ be the bin width. For each bin index $b\in\{0,\ldots,B-1\}$, we define the corresponding set of pixels as:
\begin{equation}
\mathcal{R}_b =
\left\{
i\in\mathcal{R}
\;\middle|\;
b\Delta
\le
\phi_i
<
(b+1)\Delta
\right\}.
\end{equation}

The value of the $b$-th bin is computed by accumulating the gradient magnitudes of pixels assigned to that bin:
\begin{equation}
h_b =
\sum_{i\in\mathcal{R}_b}
m_i .
\end{equation}

We identify the dominant bin and define the center of its interval as the coarse
center orientation:
\begin{equation}
b^\ast =
\arg\max_{b}
h_b,
\qquad
\theta_{\text{c}} =
\left(
b^\ast+\frac{1}{2}
\right)
\Delta .
\end{equation}

Using this coarse center as an anchor, we collect retained pixels whose
edge-tangent orientations lie within one bin width of $\theta_{\text{c}}$:
\begin{equation}
\mathcal{N}
=
\left\{
i\in\mathcal{R}
\;\middle|\;
\left|
\mathcal{W}_{\pi}
\left(
\phi_i-\theta_{\text{c}}
\right)
\right|
\le
\Delta
\right\},
\end{equation}
where $\mathcal{W}_{\pi}(\cdot)$ denotes a $\pi$-periodic wrapping function that maps an angle to the orientation range $[-\pi/2,\pi/2)$. The pseudo-angle is computed as a magnitude-weighted $\pi$-periodic circular mean over the selected pixels:
\begin{equation}
\begin{aligned}
S
&=
\sum_{i\in\mathcal{N}}
m_i
\sin
\left(
2\phi_i
\right),
\\
C
&=
\sum_{i\in\mathcal{N}}
m_i
\cos
\left(
2\phi_i
\right),
\\
\theta_{\text{grad}}
&=
\mathcal{W}_{\pi}
\left[
\frac{1}{2}
\operatorname{atan2}
\left(
S,
C
\right)
\right].
\end{aligned}
\end{equation}
This formulation accounts for the $\pi$-periodicity of orientation, treating $\theta$ and $\theta+\pi$ as the same orientation.

We further estimate the reliability of the pseudo-angle from the concentration of the orientation histogram. The normalized histogram is given by:
\begin{equation}
p_b =
\frac{
h_b
}{
\sum_{j=0}^{B-1}
h_j
}.
\end{equation}
The confidence score is defined as the complement of its normalized entropy:
\begin{equation}
q_{\text{grad}}
=
1+
\frac{
\sum_{b=0}^{B-1}
p_b
\log p_b
}{
\log B
}.
\end{equation}
The resulting $\theta_{\text{grad}}$ and $q_{\text{grad}}$ are used as an
object-specific orientation target and its confidence weight, respectively.

\subsection{Gradient-Guided Orientation Supervision}
Given the gradient-derived pseudo-angle $\theta_{\text{grad}}$, we transform it
according to each training view so that the target is expressed in the same
coordinate frame as the prediction. For $v\in\{\mathrm{ori},\mathrm{rot},\mathrm{flp}\}$,
the view-specific pseudo-angle target is defined as:
\begin{equation}
\tilde{\theta}_{\text{grad}}^{v}
=
\begin{cases}
\theta_{\text{grad}},
& v=\mathrm{ori}, \\[2pt]
\mathcal{W}_{\pi}
\left(
\theta_{\text{grad}}+R
\right),
& v=\mathrm{rot}, \\[2pt]
\mathcal{W}_{\pi}
\left(
-\theta_{\text{grad}}
\right),
& v=\mathrm{flp},
\end{cases}
\end{equation}
where $R$ denotes the rotation angle applied to the rotated view. The confidence
score $q_{\text{grad}}$ is shared across the three view-specific targets.

For each prediction, we compute the $\pi$-periodic angular discrepancy between
the predicted angle and the corresponding pseudo-angle target:
\begin{equation}
\delta
=
\left|
\mathcal{W}_{\pi}
\left(
\theta_{\text{pred}}^{v}
-
\tilde{\theta}_{\text{grad}}^{v}
\right)
\right|,
\qquad
\delta\in\left[0,\frac{\pi}{2}\right].
\end{equation}
Since local gradients may indicate either object axis, a pseudo-angle shifted by $\pi/2$ remains valid. We therefore relax the angular discrepancy as:
\begin{equation}
d_{\text{grad}}
=
\min
\left(
\delta,
\frac{\pi}{2}-\delta
\right),
\qquad
d_{\text{grad}}\in\left[0,\frac{\pi}{4}\right].
\end{equation}
This relaxation treats orthogonal object axes as equivalent for gradient-based
supervision, while box regression handles the width-height assignment.

We apply the gradient-guided loss only to predictions whose relaxed angular
distance is within the threshold $\tau_{\text{grad}}$. Let $\mathcal{V}$ denote
the set of valid assigned predictions:
\begin{equation}
\mathcal{V}
=
\left\{
n
\;\middle|\;
d_{\text{grad}}^{(n)}
\le
\tau_{\text{grad}}
\right\}.
\end{equation}
The gradient-guided loss is then defined as:
\begin{equation}
\mathcal{L}_{\text{grad}}
=
\frac{1}{|\mathcal{P}|}
\sum_{n\in\mathcal{V}}
q_{\text{grad}}^{(n)}
\,
\mathrm{smooth}_{L1}
\left(
d_{\text{grad}}^{(n)}\right),
\end{equation}
where $\mathcal{P}$ denotes the set of positive
predictions. For each prediction, the pseudo-angle target and confidence score are obtained
from its matched GT instance. The confidence score weights the reliability of
the gradient-derived pseudo-angle, and predictions outside the angular threshold are excluded from the gradient-guided supervision.

\subsection{Loss Functions}
\label{subsec:loss_function}

Following multi-view HBox-supervised OOD methods~\cite{yu2023h2rbox,lee2025abbspo}, we regularize angle predictions under rotation and vertical flipping. Let $\theta$, $\theta_{\text{rot}}$, and $\theta_{\text{flp}}$ denote the predicted angles from the original, rotated, and flipped views, respectively. The rotation- and flip-consistency losses are defined as:
\begin{equation}
\mathcal{L}_{\text{rot}} = \ell_s(\theta_{\text{rot}} - \theta, R),
\quad
\mathcal{L}_{\text{flp}} = \ell_s(\theta_{\text{flp}} + \theta, 0),
\end{equation}
where $R$ is the rotation angle applied to the rotated view, and $\ell_s$ denotes the Smooth L1-based snap loss~\cite{yu2023h2rbox}.

The multi-view angle consistency loss is then given by:
\begin{equation}
\mathcal{L}_{\text{ang}}
=
\lambda_r \mathcal{L}_{\text{rot}}
+
\lambda_f \mathcal{L}_{\text{flp}}.
\end{equation}

While $\mathcal{L}_{\text{ang}}$ enforces view-consistent angle predictions under geometric transformations, $\mathcal{L}_{\text{grad}}$ provides object-level orientation supervision estimated from local gradients.

The baseline training objective is defined as:
\begin{equation} \mathcal{L}_{\text{base}} = \lambda_{\text{ang}}\mathcal{L}_{\text{ang}} + \lambda_{\text{reg}}\mathcal{L}_{\text{reg}} + \lambda_{\text{cn}}\mathcal{L}_{\text{cn}} + \lambda_{\text{cls}}\mathcal{L}_{\text{cls}} ,
\end{equation}
where $\mathcal{L}_{\text{reg}}$, $\mathcal{L}_{\text{cn}}$, and $\mathcal{L}_{\text{cls}}$ denote the IoU-based regression loss~\cite{lee2025abbspo}, centerness loss, and focal loss-based classification loss~\cite{yu2023h2rbox}, respectively. The overall objective is defined as \begin{equation} \mathcal{L}_{\text{total}} = \mathcal{L}_{\text{base}} + \lambda_{\text{grad}}\mathcal{L}_{\text{grad}} . \end{equation}

\begin{table*}[!t]
\centering
\scriptsize
\setlength{\tabcolsep}{2.2pt}
\renewcommand{\arraystretch}{1.05}
\resizebox{\textwidth}{!}{
\begin{tabular}{l|l|cccccccccccccccccccc|c}
\toprule
 & Methods 
 & APL & APO & BF & BC & BR & CH & ESA & ETS & DAM & GF 
 & GTF & HA & OP & SH & STA & STO & TC & TS & VE & WM
 & AP$_{50}$ \\
\midrule

\multirow{6}{*}{\rotatebox[origin=c]{90}{$\mathcal{S}_{R}$}}
& RetinaNet~\cite{lin2017focal}
& 59.8 & 19.3 & 69.7 & 81.3 & 17.2 & 72.7 & 68.7 & 49.4 & 18.4 & 69.5
& 71.3 & 33.3 & 34.1 & 75.8 & 67.1 & 59.6 & 81.0 & 44.1 & 38.0 & 62.5
& 54.64 \\

& FCOS~\cite{tian2019fcos}
& 62.1 & 37.9 & 74.6 & 81.2 & 32.9 & 72.1 & 75.3 & 61.8 & 27.4 & 69.1
& 78.7 & 34.4 & 50.6 & 80.1 & 68.6 & 68.1 & 81.3 & 49.1 & 43.4 & 64.5
& 60.66 \\

& Oriented R-CNN~\cite{xie2021oriented}
& 63.0 & 36.7 & 71.9 & 81.6 & 41.1 & 72.6 & 77.8 & 65.5 & 24.8 & 72.9
& 82.1 & 40.9 & 56.5 & 81.2 & 73.4 & 62.4 & 81.5 & 53.3 & 43.3 & 65.6
& \textbf{62.41} \\

& GWD~\cite{yang2021rethinking} (RetinaNet)
& 61.5 & 23.6 & 73.6 & 81.1 & 17.4 & 72.7 & 68.3 & 47.2 & 20.7 & 71.2
& 73.2 & 33.9 & 34.3 & 77.6 & 64.7 & 57.5 & 80.9 & 42.1 & 39.7 & 60.2
& 55.07 \\

& KLD~\cite{yang2021learning} (RetinaNet)
& 57.8 & 22.6 & 71.5 & 81.2 & 16.9 & 72.7 & 68.9 & 52.1 & 20.6 & 73.5
& 71.0 & 33.7 & 33.2 & 77.1 & 68.9 & 59.9 & 80.9 & 43.9 & 39.1 & 60.9
& 55.32 \\

& KFIoU~\cite{yang2022kfiou} (RetinaNet)
& 60.6 & 36.6 & 73.6 & 80.9 & 27.0 & 72.6 & 73.4 & 56.5 & 25.4 & 73.9
& 72.0 & 32.9 & 45.8 & 75.8 & 65.2 & 57.6 & 80.0 & 48.0 & 40.1 & 58.8
& 57.84 \\

\midrule
\multirow{4}{*}{\rotatebox[origin=c]{90}{$\mathcal{S}_{H}$}}
& H2RBox~\cite{yang2022h2rbox}
& 57.1 & 14.4 & 72.2 & 82.6 & 17.5 & 71.2 & 56.5 & 55.2 & 14.0 & 67.7
& 77.9 & 31.0 & 40.7 & 76.3 & 66.2 & 63.4 & \textbf{81.5} & 50.4 & 38.0 & 57.6
& 54.57 \\

& H2RBox-v2~\cite{yu2023h2rbox}
& 55.5 & 17.8 & \textbf{76.9} & 80.5 & 27.7 & \textbf{72.2} & 63.0 & 58.6 & 24.4 & \textbf{73.9}
& 80.3 & 33.9 & 47.2 & 77.4 & 58.7 & 60.9 & 81.4 & 48.1 & 41.1 & 53.9
& 56.67 \\

& EIE-Det*$^\dagger$~\cite{eie}
& 65.2 & 29.4 & 68.5 & 83.1 & 26.6 & \textbf{72.2} & 72.0 & 62.9 & 23.7 & 67.6
& 77.1 & 32.5 & 45.5 & \textbf{79.8} & 60.2 & \textbf{67.1} & 81.3 & 51.9 & 42.7 & \textbf{63.5}
& 58.64 \\

& ABBSPO$^\dagger$~\cite{lee2025abbspo}
& 66.4 & 28.9 & 76.4 & 85.6 & \textbf{32.5} & 72.1 & \textbf{73.5} & 59.9 & 23.7 & 70.7
& 81.0 & 35.2 & 49.9 & 78.6 & 66.7 & 66.1 & 81.4 & 49.1 & 41.2 & 62.1
& 60.10 \\

\rowcolor{gray!15}
& GradOrient (Ours)
& \textbf{67.8} & \textbf{37.0} & 76.1 & \textbf{87.1} & 32.0 & 72.1 & 73.1 & \textbf{64.1} & \textbf{26.5} & 72.0
& \textbf{81.4} & \textbf{35.9} & \textbf{51.9} & 79.4 & \textbf{67.1} & 67.0 & \textbf{81.5} & \textbf{53.1} & \textbf{43.1} & 62.3
& \textbf{61.50} \\

\bottomrule
\end{tabular}
}
\vspace{-0.2cm}
\caption{Results of each category on the DIOR-R~\cite{cheng2022anchor} for RBox-supervised ($\mathcal{S}_{R}$) and HBox-supervised ($\mathcal{S}_{H}$). The performance of the compared methods is reported from a previous study \cite{lee2025abbspo}. * indicates our own implementation, and $^\dagger$ denotes models trained in this work.}
\label{tab:dior_r_results}
\vspace{-0.1cm}
\end{table*}

\begin{table*}[t]
\centering
\renewcommand{\arraystretch}{0.9}
\setlength{\tabcolsep}{6pt} 
\resizebox{\textwidth}{!}{
\begin{tabular}{l|l|ccccccccccccccc|c}
\toprule
 & Methods 
 & PL & BD & BR & GTF & SV & LV & SH & TC & BC & ST & SBF & RA & HA & SP & HC
 & AP$_{50}$ \\
\midrule

\multirow{7}{*}{\rotatebox[origin=c]{90}{$\mathcal{S}_{R}$}}
& RetinaNet~\cite{lin2017focal}
& 87.5 & 75.1 & 39.9 & 59.6 & 66.3 & 66.3 & 78.2 & 90.5 & 55.0 & 62.7 & 47.1 & 63.6 & 59.4 & 55.1 & 43.0
& 63.3 \\

& FCOS~\cite{tian2019fcos}
& 88.8 & 74.0 & 46.8 & 59.1 & 70.1 & 81.4 & 87.7 & 90.7 & 67.7 & 68.3 & 60.2 & 66.1 & 64.9 & 58.7 & 44.0
& 68.6 \\

& Oriented R-CNN~\cite{xie2021oriented}
& 89.3 & 76.1 & 53.8 & 78.7 & 68.6 & 84.9 & 89.3 & 90.8 & 74.3 & 62.8 & 66.3 & 66.5 & 74.7 & 58.6 & 46.8
& \textbf{72.1} \\

& Oriented RepPoints~\cite{li2022oriented}
& 89.7 & 80.1 & 50.5 & 74.4 & 75.0 & 82.0 & 88.7 & 90.4 & 64.0 & 70.0 & 45.7 & 60.6 & 73.6 & 60.4 & 42.8
& 69.86 \\

& GWD~\cite{yang2021rethinking} (RetinaNet)
& 88.2 & 74.9 & 41.3 & 60.5 & 66.7 & 68.1 & 85.8 & 90.5 & 50.4 & 66.8 & 45.8 & 65.1 & 60.7 & 52.9 & 38.9
& 63.77 \\

& KLD~\cite{yang2021learning} (RetinaNet)
& 88.4 & 75.8 & 41.4 & 60.0 & 66.1 & 68.8 & 84.7 & 90.6 & 56.8 & 60.4 & 50.4 & 70.1 & 60.0 & 50.5 & 45.7
& 64.65 \\

& KFIoU~\cite{yang2022kfiou} (RetinaNet)
& 84.4 & 74.3 & 40.7 & 55.2 & 57.9 & 56.9 & 76.4 & 71.2 & 46.1 & 64.8 & 54.3 & 65.0 & 58.3 & 48.7 & 42.9
& 59.81 \\

\midrule
\multirow{4}{*}{\rotatebox[origin=c]{90}{$\mathcal{S}_{H}$}}
& H2RBox~\cite{yang2022h2rbox}
& \textbf{89.5} & 73.1 & 37.3 & 55.1 & 70.7 & 76.4 & 85.4 & 90.3 & 66.5 & 67.3 & 59.6 & 64.9 & 60.6 & \textbf{57.9} & 36.5
& 66.07 \\

& H2RBox-v2~\cite{yu2023h2rbox}
& 89.4 & 74.8 & 45.4 & 56.0 & 70.3 & 76.6 & 87.9 & \textbf{90.5} & 69.3 & 67.5 & 56.7 & 64.7 & 65.3 & 55.5 & 45.5
& 67.69 \\

& EIE-Det*$^\dagger$~\cite{eie}
& 89.1 & 70.8 & 44.5 & 56.0 & 70.1 & 74.8 & 84.2 & 90.4
& 61.7 & 66.5 & 56.0 & 61.4 & 62.3 & 51.6 & 46.8
& 65.75 \\

& ABBSPO$^\dagger$~\cite{lee2025abbspo}
& 89.1 & \textbf{75.7} & \textbf{47.3} & \textbf{58.1} & 69.9 & 77.8 & 88.1 & \textbf{90.5} & 69.3 & 67.8 & 67.1 & 64.3 & \textbf{71.8} & 55.3 & 44.9
& 69.13 \\

\rowcolor{gray!15}
& GradOrient (Ours)
& 89.4 & 75.2 & 46.6 & 58.0 & \textbf{71.1} & \textbf{77.9} & \textbf{88.2} & \textbf{90.5} & \textbf{72.0} & \textbf{68.6} & \textbf{67.3} & \textbf{65.7} & 71.5 & 56.4 & \textbf{56.0}
& \textbf{70.29} \\

\bottomrule
\end{tabular}
}
\vspace{-0.2cm}
\caption{Results of each category on the DOTA-v1.0~\cite{xia2018dota} for RBox-supervised ($\mathcal{S}_{R}$) and HBox-supervised ($\mathcal{S}_{H}$). Results for the compared methods are taken from a previous study \cite{lee2025abbspo}. * indicates our own implementation, and $^\dagger$ denotes models trained in this work.}
\label{tab:dota_results}
\vspace{0.0cm}
\end{table*}

\section{Orientation-aware Evaluation Metric}
\label{subsec:orientation_aware_metric}

\begin{figure}[tb]
    \centering    \includegraphics[width=\columnwidth]{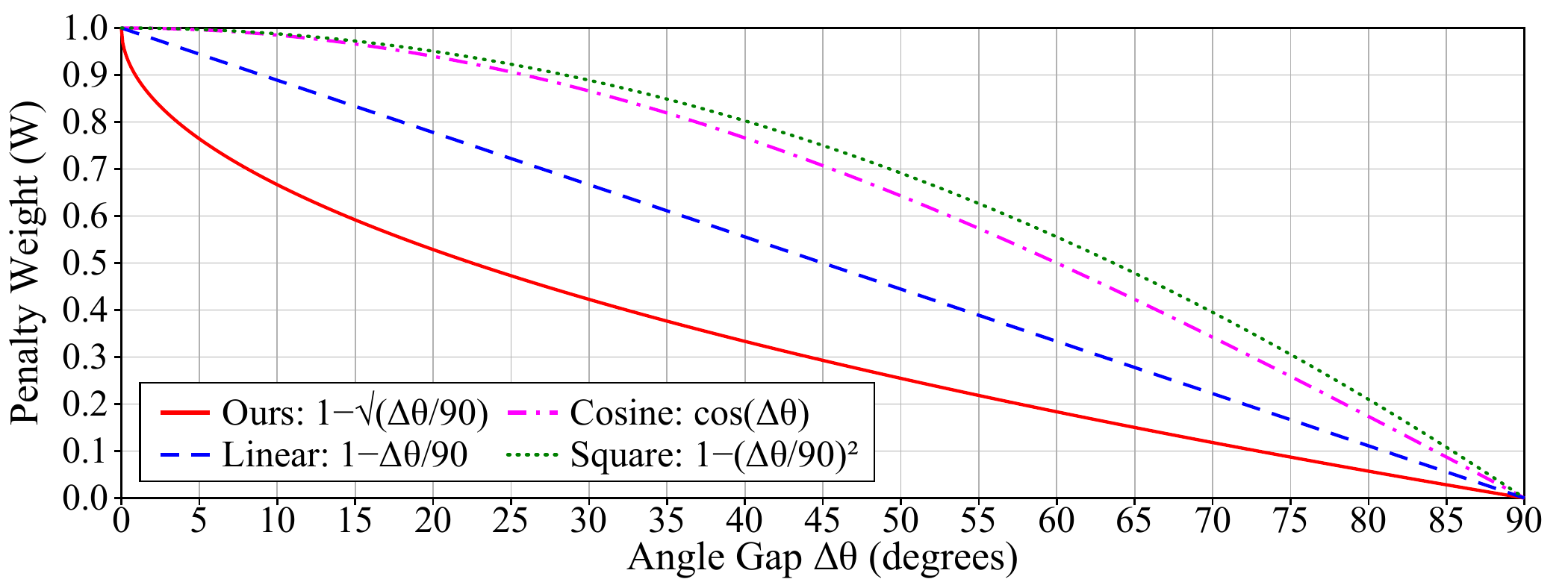}
    \vspace{-0.5cm}
    \caption{Comparison of angular penalty weights $W$. We adopt the square-root penalty, which imposes a stronger penalty for small angular errors than cosine, linear, and square penalties.}
    \label{fig:comparison}
    \vspace{-0.4cm}
\end{figure}

The standard average precision (AP) metric does not explicitly account for angular errors. 
An OBB prediction with a significantly misaligned angle can still be counted as a true positive (TP) as long as its spatial intersection over union (IoU) satisfies the threshold. To address this issue, we propose orientation-aware AP (OAP), which incorporates an angular penalty into the standard AP evaluation procedure.

To measure the angular discrepancy consistently, we extract the predicted and ground-truth (GT) orientations, denoted by $\theta_{\text{pred}}$ and $\theta_{\text{gt}}$, respectively, under the long-edge 90° (\textit{le90}) representation. Since oriented boxes exhibit periodic angular representations, we compute the acute angular difference $\Delta\theta\in[0,\pi/2]$ as follows:
\begin{equation}
\label{eq:angle_gap}
    \Delta\theta = \min(|\theta_{\text{pred}} - \theta_{\text{gt}}|, 
    \pi - |\theta_{\text{pred}} - \theta_{\text{gt}}|).
\end{equation}

We then define an angular penalty weight $W$ as:
\begin{equation}
\label{eq:angle_weight}
    W = 1 - \sqrt{\frac{\Delta\theta}{\theta_{\max}}},
\end{equation}
where $\theta_{\max}=\pi/2$ denotes the maximum possible angular deviation. The weight $W\in[0,1]$ represents the degree of orientation agreement between the predicted and GT boxes. When the two boxes have the same orientation, $\Delta\theta=0$ and thus $W=1$, indicating that no angular penalty is applied.
As the angular error increases, $W$ decreases, and it becomes $0$ when the angular deviation reaches $\theta_{\max}$.

\cref{fig:comparison} compares several choices for the angular penalty function, including cosine, linear, square, and square-root forms. 
We adopt the square-root penalty because it decreases more rapidly than the other alternatives for small angular errors. 
This behavior is desirable for orientation-aware evaluation, since even a moderate angular mismatch can indicate inaccurate orientation estimation, especially when the spatial IoU remains high. 
Therefore, the proposed penalty enforces stricter orientation consistency while still preserving the standard IoU-based localization criterion.

Using this angular weight, we define the Orientation-Aware IoU (OIoU) as $W \times \mathrm{IoU}$. During evaluation, a prediction is considered a TP only when its OIoU exceeds a predefined threshold (e.g., 0.5 for OAP$_{50}$). 
The standard AP calculation is then applied using these orientation-aware matches, yielding the final OAP score. 
As a result, OAP preserves the conventional AP evaluation pipeline while penalizing predictions with inaccurate orientations.

\begin{table*}[tb]
\centering
\scriptsize
\setlength{\tabcolsep}{1.8pt}
\renewcommand{\arraystretch}{1.05}
\resizebox{\textwidth}{!}{
\begin{tabular}{l|l|cccccccccccccc|c|ccccc|c}
\toprule
 & \multirow{2}{*}{Methods}
 & \multicolumn{15}{c|}{OAP$_{50}$}
 & \multicolumn{5}{c|}{AngleAcc$_\theta$}
 & \multirow{2}{*}{AngleGap} \\
\noalign{\vskip 1pt}
\cline{3-17} \cline{18-22}
\noalign{\vskip 2pt}
 & 
 & APL & APO & BC & BR & ESA & ETS & DAM & GTF & HA & OP & SH & TC & TS & VE & ALL
 & 1$^\circ$ & 2$^\circ$ & 3$^\circ$ & 4$^\circ$ & 5$^\circ$
 & \\
\midrule

\multirow{6}{*}{\rotatebox[origin=c]{90}{$\mathcal{S}_{R}$}}
& RetinaNet~\cite{lin2017focal}
& 22.4 & 6.3 & 79.5 & 10.3 & 54.7 & 41.2 & 7.0 & 69.2 & 23.9 & 28.3 & 68.1 & 80.7 & 34.3 & 30.4 & 39.7
& 39.84 & 60.15 & 71.36 & 78.54 & 83.29 & 5.263 \\

& FCOS~\cite{tian2019fcos}
& 24.9 & 17.6 & 79.6 & 24.9 & 50.2 & 48.8 & 10.2 & 71.7 & 17.4 & 42.0 & 69.0 & 81.1 & 33.0 & 32.2 & 43.0
& 28.70 & 49.16 & 62.80 & 71.90 & 77.95 & 5.317 \\

& Oriented R-CNN~\cite{xie2021oriented}
& 40.6 & 21.0 & 81.2 & 34.3 & 65.3 & 59.3 & 17.5 & 80.6 & 32.9 & 49.0 & 80.7 & 81.4 & 44.9 & 41.1 & \textbf{52.1}
& \textbf{43.67} & \textbf{64.98} & \textbf{77.20} & \textbf{84.68} & \textbf{89.33} & \textbf{3.262} \\

& GWD~\cite{yang2021rethinking} (RetinaNet)
& 13.8 & 10.5 & 77.3 & 9.7 & 50.7 & 37.9 & 3.9 & 68.3 & 22.8 & 27.1 & 68.5 & 80.7 & 32.6 & 29.8 & 38.1
& 31.30 & 52.23 & 65.77 & 74.24 & 79.56 & 5.911 \\

& KLD~\cite{yang2021learning} (RetinaNet)
& 18.2 & 11.3 & 77.6 & 12.1 & 52.6 & 39.7 & 7.4 & 69.0 & 25.4 & 30.2 & 68.9 & 80.7 & 32.9 & 30.9 & 39.8
& 33.16 & 55.11 & 68.27 & 76.41 & 81.47 & 5.567 \\

& KFIoU~\cite{yang2022kfiou} (RetinaNet)
& 17.9 & 17.9 & 70.1 & 18.1 & 49.0 & 37.7 & 8.2 & 60.6 & 24.8 & 37.9 & 65.6 & 71.1 & 36.5 & 29.8 & 38.9
& 29.88 & 51.08 & 64.37 & 72.96 & 78.71 & 6.126 \\

\midrule
\multirow{4}{*}{\rotatebox[origin=c]{90}{$\mathcal{S}_{H}$}}
& H2RBox~\cite{yang2022h2rbox}
& 9.7 & 0.9 & 80.5 & 14.9 & 36.1 & 46.0 & 10.5 & 69.5 & 24.1 & 33.1 & 66.8 & 81.3 & 36.8 & 28.5 & 38.5
& 33.25 & 52.52 & 64.36 & 72.28 & 77.60 & 5.387 \\

& H2RBox-v2~\cite{yu2023h2rbox}
& 22.6 & 8.0 & 80.5 & \textbf{26.2} & 34.2 & 43.8 & 13.9 & 74.0 & 26.0 & 43.1 & 66.7 & 81.3 & 33.5 & 33.4 & 42.0
& 29.75 & 51.74 & 66.63 & 76.18 & 82.57 & 4.624 \\

& EIE-Det*~\cite{eie}
& 27.5 & 13.8 & 80.5 & 18.9 & \textbf{52.7} & 50.6 & 7.9 & 71.6 & 24.1 & 36.8
& \textbf{70.0} & 81.3 & \textbf{41.2} & 33.0 & 43.6
& \textbf{40.14} & \textbf{59.82} & 71.23 & 78.57 & 83.40 & 4.393 \\

& ABBSPO~\cite{lee2025abbspo}
& 27.4 & 10.4 & 80.1 & 24.7 & 40.3 & 41.8 & 9.8 & 74.0 & 24.2 & 39.5 & 67.7 & 81.2 & 31.8 & 32.4 & 41.8
& 30.99 & 51.80 & 66.06 & 75.36 & 81.52 & 5.019 \\

\rowcolor{gray!15}
& GradOrient (Ours)
& \textbf{33.6} & \textbf{20.2} & \textbf{80.6} & 25.5 & 51.6 & \textbf{50.8} & \textbf{14.4} & \textbf{74.5} & \textbf{26.6} & \textbf{44.0} & 69.3 & \textbf{81.4} & 39.5 & \textbf{33.5} & \textbf{46.1}
& 35.70 & 59.43 & \textbf{74.15} & \textbf{82.91} & \textbf{88.14} & \textbf{4.159} \\

\bottomrule
\end{tabular}
}
\vspace{-0.2cm}
\caption{Orientation-aware results on the DIOR-R~\cite{cheng2022anchor} for RBox-supervised ($\mathcal{S}_{R}$) and HBox-supervised ($\mathcal{S}_{H}$). AngleAcc$_\theta$ reports the percentage of AP$_{50}$-matched TP pairs whose angular error is within each threshold. * indicates our own implementation.}
\label{tab:dior_r_orientation_results}
\vspace{-0.0cm}
\end{table*}

\begin{table*}[tb]
\centering
\scriptsize
\setlength{\tabcolsep}{2.2pt}
\renewcommand{\arraystretch}{1.05}
\resizebox{\textwidth}{!}{
\begin{tabular}{l|l|cccccccccccc|c|ccccc|c}
\toprule
 & \multirow{2}{*}{Methods}
 & \multicolumn{13}{c|}{OAP$_{50}$}
 & \multicolumn{5}{c|}{AngleAcc$_\theta$}
 & \multirow{2}{*}{AngleGap} \\
\noalign{\vskip 1pt}
\cline{3-15} \cline{16-20}
\noalign{\vskip 2pt}
 & 
 & PL & BR & GTF & SV & LV & SH & TC & BC & SBF & HA & SP & HC & ALL
 & 1$^\circ$ & 2$^\circ$ & 3$^\circ$ & 4$^\circ$ & 5$^\circ$
 & \\
\midrule

\multirow{7}{*}{\rotatebox[origin=c]{90}{$\mathcal{S}_{R}$}}
& RetinaNet~\cite{lin2017focal}
& 61.0 & 24.8 & 48.8 & 59.5 & 60.7 & 73.4 & 90.3 & 46.9 & 41.8 & 38.8 & 20.8 & 7.0 & 47.8
& 32.14 & 55.93 & 72.02 & 81.86 & 87.80 & 4.106 \\

& FCOS~\cite{tian2019fcos}
& 63.1 & 26.8 & 42.8 & 60.6 & 75.7 & 77.1 & 90.3 & 54.4 & 39.8 & 38.7 & 23.5 & 16.0 & 50.7
& 26.32 & 47.64 & 63.00 & 73.80 & 81.13 & 4.349 \\

& Oriented R-CNN~\cite{xie2021oriented}
& 73.6 & 37.9 & 72.9 & 62.9 & 83.7 & 88.5 & 90.7 & 72.0 & 60.8 & 63.3 & 24.5 & 17.8 & \textbf{62.4}
& \textbf{36.83} & \textbf{61.88} & \textbf{77.04} & \textbf{86.10} & \textbf{91.26} & \textbf{3.035} \\

& Oriented RepPoints~\cite{li2022oriented}
& 74.2 & 31.9 & 64.5 & 66.2 & 78.7 & 87.3 & 90.0 & 53.1 & 41.0 & 52.0 & 25.8 & 9.1 & 56.1
& 33.80 & 58.84 & 74.66 & 84.40 & 90.03 & 3.304 \\

& GWD~\cite{yang2021rethinking} (RetinaNet)
& 52.5 & 20.1 & 50.7 & 60.9 & 63.6 & 77.3 & 90.4 & 47.9 & 33.3 & 37.0 & 21.5 & 9.1 & 47.0
& 29.17 & 52.23 & 68.01 & 78.12 & 84.58 & 4.175 \\

& KLD~\cite{yang2021learning} (RetinaNet)
& 49.9 & 17.8 & 52.3 & 58.2 & 61.5 & 76.0 & 90.2 & 45.5 & 35.5 & 32.4 & 20.3 & 8.7 & 45.7
& 27.92 & 49.82 & 65.44 & 75.67 & 82.22 & 4.629 \\

& KFIoU~\cite{yang2022kfiou} (RetinaNet)
& 32.4 & 17.8 & 43.2 & 51.0 & 46.0 & 65.6 & 62.6 & 37.4 & 26.0 & 34.4 & 18.8 & 16.3 & 37.6
& 26.40 & 48.27 & 64.02 & 74.84 & 81.73 & 5.358 \\

\midrule
\multirow{4}{*}{\rotatebox[origin=c]{90}{$\mathcal{S}_{H}$}}
& H2RBox~\cite{yang2022h2rbox}
& 58.0 & 17.4 & 37.6 & 59.4 & 65.4 & 77.3 & \textbf{90.4} & 55.9 & 45.0 & 29.4 & 20.5 & 11.4 & 47.3
& 28.22 & 49.86 & 65.41 & 76.19 & 83.17 & 4.360 \\

& H2RBox-v2~\cite{yu2023h2rbox}
& 63.4 & 29.7 & 45.3 & 60.7 & 73.5 & 78.5 & 90.3 & 65.5 & 49.8 & 41.8 & 23.6 & 16.2 & 53.2
& 31.66 & 55.52 & 71.32 & 81.60 & 87.73 & 3.752 \\

& EIE-Det*~\cite{eie}
& 58.4 & 23.1 & \textbf{46.6} & 61.2 & 66.3 & 76.9
& 90.3 & 55.5 & 35.7 & 30.4 & 20.6 & \textbf{24.7}
& 49.1
& 27.10 & 49.37 & 65.88 & 76.86 & 84.10
& 4.415 \\

& ABBSPO~\cite{lee2025abbspo}
& 64.0 & 28.9 & 45.0 & 61.4 & 74.3 & 79.0 & \textbf{90.4} & 66.8 & 51.3 & 46.6 & 23.9 & 23.9 & 54.6
& 31.50 & 55.68 & 71.97 & 81.85 & 87.99 & 3.676 \\

\rowcolor{gray!15}
& GradOrient (Ours)
& \textbf{64.9} & \textbf{30.4} & 42.5 & \textbf{63.8} & \textbf{76.4} & \textbf{79.1} & \textbf{90.4} & \textbf{70.9} & \textbf{55.6} & \textbf{50.7} & \textbf{25.9} & 11.9 & \textbf{55.2}
& \textbf{35.91} & \textbf{62.16} & \textbf{77.62} & \textbf{86.57} & \textbf{91.30} & \textbf{3.348} \\

\bottomrule
\end{tabular}
}
\vspace{-0.2cm}
\caption{Orientation-aware results on the DOTA-v1.0~\cite{xia2018dota} for RBox-supervised ($\mathcal{S}_{R}$) and HBox-supervised ($\mathcal{S}_{H}$). AngleAcc$_\theta$ reports the percentage of AP$_{50}$-matched TP pairs whose angular error is within each threshold. * indicates our own implementation.}
\label{tab:dota_orientation_results}
\vspace{0.0cm}
\end{table*}

\section{Experiments}
\label{sec:experiments}

\subsection{Setup}

\textbf{Datasets.}
\label{subsec:datasets}
We utilize two prominent benchmarks for aerial object detection: DOTA-v1.0 \cite{xia2018dota} and DIOR-R \cite{cheng2022anchor}. DOTA-v1.0 is a comprehensive dataset comprising 2,806 images with resolutions between 800 and 4,000 pixels, featuring 188,282 oriented bounding box annotations distributed across 15 classes. Similarly, DIOR-R extends the original DIOR dataset with oriented labels, offering 23,463 remote sensing images and 192,518 instances across 20 categories.

\noindent\textbf{Implementation Details.}
Our method is implemented based on the ABBSPO \cite{lee2025abbspo} framework, and all models are trained for 12 epochs under the same network setup. All experiments are conducted using an NVIDIA RTX 4090 GPU. For DOTA-v1.0, we train on the training set and evaluate on the validation set, as our angle-based metric requires access to ground-truth annotations. We use $\tau_{\text{grad}}=6^\circ$ for DIOR-R and $4^\circ$ for DOTA-v1.0. We fix the loss weights to $\lambda_{r}=1.0$, $\lambda_{f}=0.05$, $\lambda_{\text{ang}}=0.6$, $\lambda_{\text{reg}}=\lambda_{\text{cn}}=\lambda_{\text{cls}}=1.0$, and $\lambda_{\text{grad}}=10.0$, with $\beta=0.1$ for the Smooth $L_1$ loss.

\noindent\textbf{Metrics Used in Experiments.}
We report four evaluation metrics: AP$_{50}$, OAP$_{50}$, AngleGap, and AngleAcc. All angular quantities are computed under the \textit{le90} convention. AP$_{50}$ follows the standard detection protocol, where a prediction is considered a TP when its IoU with the matched GT box is at least 0.5. OAP$_{50}$ follows the same AP calculation procedure but uses OIoU instead of IoU for matching, thereby accounting for both spatial overlap and orientation consistency. AngleGap measures the average angular difference between the predicted and GT boxes among TP pairs matched under AP$_{50}$, where a lower value indicates more accurate orientation estimation. AngleAcc$_{k^\circ}$ denotes the percentage of AP$_{50}$-matched TP pairs whose angular error is within $k^\circ$, and we report AngleAcc from $1^\circ$ to $5^\circ$.

\subsection{Results}
\label{subsec:experimental_results}

\textbf{Quantitative Results.}
\cref{tab:dior_r_results,tab:dota_results} compare GradOrient with existing RBox-supervised and HBox-supervised oriented object detectors under the conventional AP$_{50}$ metric. GradOrient achieves 61.50 AP$_{50}$ on DIOR-R and 70.29 AP$_{50}$ on DOTA-v1.0, outperforming the reproduced HBox-supervised baseline ABBSPO$^\dagger$ by 1.40 and 1.16 points, respectively. \cref{tab:dior_r_orientation_results,tab:dota_orientation_results} further report the orientation-aware evaluation results. GradOrient improves OAP$_{50}$ from 41.8 to 46.1 on DIOR-R and from 54.6 to 55.2 on DOTA, while reducing AngleGap from 5.019$^\circ$ to 4.159$^\circ$ and from 3.676$^\circ$ to 3.348$^\circ$, respectively. These results indicate that the proposed gradient-guided pseudo-angle supervision improves both detection accuracy and orientation estimation.

\begin{figure*}[!tb]
    \vspace{-0.2cm}
    \centering    \includegraphics[width=0.95\textwidth]{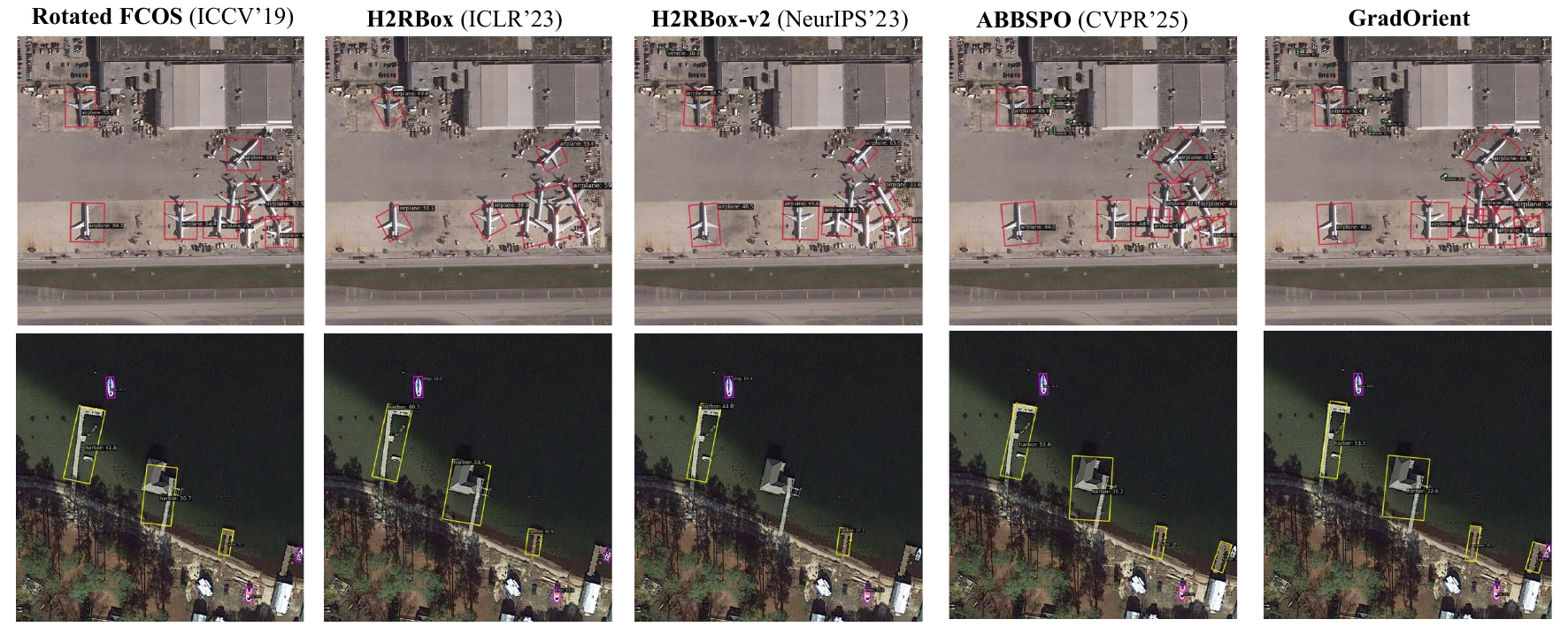}
    \vspace{-0.1cm}
    \caption{Visual comparison on DIOR-R and DOTA-v1.0. The top and bottom rows show results on DIOR-R and DOTA-v1.0, respectively.}
    \label{fig:figure4}
    \vspace{-0.0cm}

\end{figure*}

\begin{table*}[tb]
\centering
\scriptsize
\setlength{\tabcolsep}{1.2pt}
\renewcommand{\arraystretch}{1.05}

\begin{minipage}[t]{0.33\textwidth}
\centering
\resizebox{\linewidth}{!}{
\begin{tabular}{cccccccc}
\toprule
\multirow{2}{*}{$\mathcal{L}_{\text{spa}}$} & \multirow{2}{*}{$\mathcal{L}_{\text{grad}}$} 
& \multicolumn{3}{c}{DIOR-R} & \multicolumn{3}{c}{DOTA-v1.0} \\
\cmidrule(r){3-5} \cmidrule(l){6-8}
 & & AP$_{50}$ & OAP$_{50}$ & AngleGap 
   & AP$_{50}$ & OAP$_{50}$ & AngleGap \\
\midrule
 & & 59.8 & 40.2 & 5.494 & 69.3 & 54.0 & 3.915 \\ 
\checkmark & & 60.1 & 41.8 & 5.019 & 69.1 & 54.6 & 3.676 \\ 
\checkmark & \checkmark & 60.2 & 45.4 & 4.638 & 68.8 & 55.1 & 3.428 \\ 
 & \checkmark & \textbf{61.5} & \textbf{46.1} & \textbf{4.159} 
 & \textbf{70.3} & \textbf{55.2} & \textbf{3.348} \\ 
\bottomrule
\end{tabular}
}
\centerline{(a) Loss components}
\end{minipage}
\hfill
\begin{minipage}[t]{0.31\textwidth}
\centering
\resizebox{\linewidth}{!}{
\begin{tabular}{ccccccc}
\toprule
\multirow{2}{*}{top-$k$ (\%)} 
& \multicolumn{3}{c}{DIOR-R} & \multicolumn{3}{c}{DOTA-v1.0} \\
\cmidrule(r){2-4} \cmidrule(l){5-7}
 & AP$_{50}$ & OAP$_{50}$ & AngleGap 
 & AP$_{50}$ & OAP$_{50}$ & AngleGap \\
\midrule
75 & 60.4 & 44.5 & 4.549 & 68.6 & 54.7 & 3.347 \\ 
50 & 59.2 & 43.7 & 4.611 & 69.4 & 54.7 & 3.341 \\ 
25 & \textbf{61.5} & \textbf{46.1} & \textbf{4.159} 
   & \textbf{70.3} & \textbf{55.2} & 3.348 \\ 
10 & 60.3 & 45.3 & 4.563 & 68.7 & 55.0 & \textbf{3.315} \\ 
\bottomrule
\end{tabular}
}
\centerline{(b) Top-$k$ ratio}
\end{minipage}
\hfill
\begin{minipage}[t]{0.31\textwidth}
\centering
\resizebox{\linewidth}{!}{
\begin{tabular}{ccccccc}
\toprule
\multirow{2}{*}{$\lambda_{\text{grad}}$} 
& \multicolumn{3}{c}{DIOR-R} & \multicolumn{3}{c}{DOTA-v1.0} \\
\cmidrule(r){2-4} \cmidrule(l){5-7}
 & AP$_{50}$ & OAP$_{50}$ & AngleGap 
 & AP$_{50}$ & OAP$_{50}$ & AngleGap \\
\midrule
0.1 & 60.4 & 42.9 & 5.239 & 68.7 & 54.3 & 3.769 \\  
1   & 61.4 & 44.1 & 4.562 & 69.5 & \textbf{55.3} & 3.526 \\ 
10  & \textbf{61.5} & \textbf{46.1} & 4.159 
    & \textbf{70.3} & 55.2 & \textbf{3.348} \\ 
20  & 60.9 & 45.7 & \textbf{4.150} & 68.3 & 53.9 & 3.352 \\ 
\bottomrule
\end{tabular}
}
\centerline{(c) $\lambda_{\text{grad}}$}
\end{minipage}
\vspace{-0.15cm}
\caption{Ablation studies of GradOrient on DIOR-R and DOTA-v1.0. We evaluate the effects of components, top-$k$ ratio, and $\lambda_{\text{grad}}$ in terms of AP$_{50}$, OAP$_{50}$, and AngleGap. Here, $\mathcal{L}_{\text{spa}}$ denotes the symmetry-prior loss adopted from ABBSPO~\cite{lee2025abbspo}, while $\mathcal{L}_{\text{grad}}$ denotes our gradient-guided pseudo-angle loss. This indicates that gradient cues serve as a more direct image-intrinsic prior for angle learning.}

\label{tab:ablation_all}
\vspace{0.0cm}

\end{table*}

\noindent\textbf{Qualitative Results.}
\cref{fig:figure4} shows qualitative comparisons with existing oriented detectors. Compared with HBox-supervised methods, GradOrient produces rotated boxes that are more consistent with object directions, especially for elongated objects and densely arranged instances. This demonstrates that image-gradient cues provide useful orientation information that complements HBox supervision.

\begin{table}[tb]
\centering
\footnotesize 
\setlength{\tabcolsep}{2.0pt}
\begin{tabular}{ccccccc}
\toprule
\multirow{2}{*}{$\tau_{\text{grad}}~(^\circ)$} & \multicolumn{3}{c}{DIOR-R} & \multicolumn{3}{c}{DOTA-v1.0} \\
\cmidrule(r){2-4} \cmidrule(l){5-7}
 & $\text{AP}_{50}$ & $\text{OAP}_{50}$ & $\text{AngleGap}$ & $\text{AP}_{50}$ & $\text{OAP}_{50}$ & $\text{AngleGap}$ \\
\midrule
1 & 59.6 & 42.7 & 4.935 & 68.3 & 53.3 & 3.843 \\ 
2 & 61.2 & 43.4 & 4.767 & 69.2 & 54.5 & 3.554 \\ 
3 & 59.7 & 41.6 & 4.959 & 69.6 & \textbf{55.6} & 3.364 \\ 
4 & 60.7 & 44.5 & 4.568 & \textbf{70.3} & 55.2 & 3.348 \\ 
5 & 60.5 & 45.3 & 4.418 & 69.0 & 55.0 & 3.326 \\ 
6 & \textbf{61.5} & \textbf{46.1} & \textbf{4.159} & 68.4 & 54.4 & 3.302 \\ 
7 & 59.9 & 45.7 & 4.282 & 69.3 & 55.2 & \textbf{3.225} \\ 
8 & 59.4 & 44.5 & 4.271 & 68.8 & 55.4 & 3.292 \\ 
$\infty$
& 58.0 & 44.2 & 4.293
& 67.0 & 52.9 & 3.507 \\
\bottomrule
\end{tabular}
\vspace{-0.2cm}
\caption{Sensitivity analysis of the hyperparameter $\tau_{\text{grad}}$ on the DIOR-R and DOTA-v1.0 datasets.}
\label{tab:sensitivity_tau_hog}
\vspace{-0.2cm}
\end{table}

\noindent\textbf{Ablation Study.}
\cref{tab:ablation_all} analyzes the effects of the loss components, top-$k$, and $\lambda_{\text{grad}}$. Adding $\mathcal{L}_{\text{grad}}$ consistently improves OAP$_{50}$ and reduces AngleGap on both DIOR-R and DOTA-v1.0. Notably, using $\mathcal{L}_{\text{grad}}$ without $\mathcal{L}_{\text{spa}}$ achieves the best overall performance, suggesting that gradient-based pseudo-angle supervision provides a more direct and effective orientation cue than the symmetry-prior constraint, $\mathcal{L}_{\text{spa}}$ used in ABBSPO~\cite{lee2025abbspo}. This result supports our motivation that image-intrinsic gradient information can serve as a strong guidance signal for angle learning in WS-OOD. The results also show that GradOrient is relatively stable across different top-$k$ and $\lambda_{\text{grad}}$ values. In addition, \cref{tab:sensitivity_tau_hog} shows that the method remains robust to $\tau_{\text{grad}}$, while the selected thresholds provide a favorable balance between detection accuracy and orientation-aware performance.

\begin{table}[tb]
\centering
\footnotesize
\setlength{\tabcolsep}{2.0pt}

\resizebox{\columnwidth}{!}{%
\begin{tabular}{llcccccc}
\toprule
\multirow{2}{*}{Framework}
& \multirow{2}{*}{Setting}
& \multicolumn{3}{c}{DIOR-R}
& \multicolumn{3}{c}{DOTA-v1.0} \\
\cmidrule(r){3-5} \cmidrule(l){6-8}
&
& $\text{AP}_{50}$
& $\text{OAP}_{50}$
& $\text{AngleGap}$
& $\text{AP}_{50}$
& $\text{OAP}_{50}$
& $\text{AngleGap}$ \\
\midrule

\multirow{2}{*}{EIE-Det~\cite{eie}}
& Baseline
& 58.6 & 43.6 & 4.393
& 65.8 & 49.1 & 4.415 \\
& w/ $\mathcal{L}_{\mathrm{grad}}$
& 60.0 & 45.7 & 4.226 
& 67.3 & 49.6 & 4.242 \\
\midrule

\multirow{2}{*}{H2RBox \cite{yang2022h2rbox}}
& Baseline
& 54.6 & 38.5 & 5.387
& 66.1 & 47.3 & 4.360 \\
& w/ $\mathcal{L}_{\mathrm{grad}}$
& 57.5 & 41.6 & 5.007 
& 67.1 & 50.4 & 3.932 \\
\midrule

\multirow{2}{*}{H2RBox-v2 \cite{yu2023h2rbox}}
& Baseline
& 56.7 & 42.0 & 4.624
& 67.7 & 53.2 & 3.752 \\
& w/ $\mathcal{L}_{\mathrm{grad}}$
& 58.6 & 44.1 & 4.070
& 68.2 & 55.2 & 3.334 \\
\midrule

\multirow{2}{*}{Rotated FCOS \cite{tian2019fcos}}
& Baseline
& 60.7 & 43.0 & 5.317
& 68.6 & 50.7 & 4.349 \\
& w/ $\mathcal{L}_{\mathrm{grad}}$
& 60.7 & 44.8 & 5.152
& 67.5 & 51.6 & 3.959 \\
\bottomrule
\end{tabular}%
}

\vspace{-0.2cm}
\caption{Generalizability of the gradient-guided orientation supervision on DIOR-R and DOTA-v1.0.}
\label{tab:cross_framework}
\vspace{-0.4cm}
\end{table}

\noindent\textbf{Generalizability analysis.}
\cref{tab:cross_framework} evaluates the generalizability of our gradient-guided orientation supervision across both HBox-supervised~\cite{yang2022h2rbox,yu2023h2rbox,eie} and fully supervised frameworks~\cite{tian2019fcos}. We integrate our orientation supervisor into each framework and consistently observe improvements in OAP$_{50}$ of 0.5--3.1 points, together with reductions in AngleGap of 3.1--12.0\%. These results demonstrate the generalizability of our supervision across different supervision paradigms and framework designs.

\noindent\textbf{Limitations of Pseudo-angle Estimation.} Pseudo-angle estimation can become unreliable in atypical cases such as zebra-like objects, where strong internal textures produce gradients that are not aligned with the global object orientation. Similar misleading cues can also arise from cluttered backgrounds or intersecting boundaries. However, our gradient supervision is applied only when the pseudo-angle is consistent with the baseline orientation estimate through the proposed gate. Together with the original orientation supervision of the baseline, this design mitigates the influence of inaccurate pseudo-angles on orientation learning.
\section{Conclusion}
\label{sec:conclusion}
In this paper, we introduced GradOrient, a simple yet effective framework for HBox-supervised oriented object detection. Motivated by the observation that the image itself contains orientation cues, such as object boundaries and gradient patterns, GradOrient converts dominant gradient orientations within weakly annotated regions into pseudo-angle supervision, providing a direct guidance signal for angle learning without requiring RBox annotations or extra learnable parameters. Experimental results show that this gradient-guided supervision consistently improves both detection accuracy and orientation estimation quality over existing HBox-supervised baselines.

Beyond training, we also highlight that accurate angle prediction should be explicitly considered in the evaluation of oriented object detection. Conventional IoU-based AP mainly reflects spatial overlap and can overlook substantial angular errors, especially when high IoU is preserved despite inaccurate orientation. To address this limitation, we proposed orientation-aware AP (OAP), which incorporates an angular penalty into the standard AP protocol and provides a complementary measure of orientation quality. We hope this metric encourages more careful analysis of angle prediction in future OOD research.

{
    \small
    \bibliographystyle{ieeenat_fullname}
    \bibliography{main}
}

\end{document}